%% file: root.tex
\documentclass[letterpaper, 10 pt, conference]{ieeeconf}  

\IEEEoverridecommandlockouts                              

\usepackage{graphics} 
\usepackage{epsfig} 
\usepackage{mathptmx} 
\usepackage{amsmath} 
\usepackage{xcolor} 

\makeatletter
\let\NAT@parse\undefined
\makeatother
\usepackage{hyperref}

\title{\LARGE \bf
Mobile-URSONet: an Embeddable Neural Network for Onboard Spacecraft Pose Estimation
}

\author{Julien Posso{*}, Guy Bois{*}, Yvon Savaria$\dagger$
\\ Department of {*}Computer Engineering and $\dagger$Electrical Engineering, École Polytechnique de Montréal
\\ Montréal (QC), Canada
\\ Email: \{firstname.lastname\}@polymtl.ca %
}

\begin{document}

\maketitle
\thispagestyle{empty}
\pagestyle{empty}


\input{1-abstract}   
\input{2-introduction}
\input{3-methodology}
\input{4-experiments}
\input{5-conclusions}
\input{7-acknowledgment}


\bibliography{references}
\bibliographystyle{IEEEtran}

\end{document}

%% file: 1-abstract.tex
\begin{abstract}

  

Spacecraft pose estimation is an essential computer vision application that can improve the autonomy of in-orbit operations. An ESA/Stanford competition brought out solutions that seem hardly compatible with the constraints imposed on spacecraft onboard computers. URSONet is among the best in the competition for its generalization capabilities but at the cost of a tremendous number of parameters and high computational complexity. In this paper, we propose Mobile-URSONet: a spacecraft pose estimation convolutional neural network with 178 times fewer parameters while degrading accuracy by no more than four times compared to URSONet.






\end{abstract}

%% file: 2-introduction.tex
\section{INTRODUCTION}\label{sec:intro}


Estimating the relative position and orientation (commonly called Pose estimation) of a known but uncooperative spacecraft from a monocular image is an essential computer vision application that allows improving the autonomy of in-orbit spacecraft operations: formation flying, autonomous docking, satellite maintenance, debris removal, etc... \cite{proenca_deep_2020}. Debris removal is crucial for the future of low earth orbit operations as the exploitation of this orbit grows, especially with the Starlink and OneWeb constellations. A significant increase of debris will accompany these new constellations. Several research projects aim to solve this problem: RemoveDEBRIS from the Surrey Space Center, Restore-L from NASA, Phoenix program from DARPA \cite{noauthor_kelvins_nodate}, or more recently the ClearSpace-1 mission from the European Space Agency (ESA) \cite{noauthor_esa_nodate}.


Sharma \textit{et al.} were the first to propose using Convolutional Neural Networks (CNNs) for Spacecraft Pose Estimation (SPE) \cite{sharma_pose_2018}. However, the popularity of CNNs applied to SPE increased in 2019 when ESA and Stanford SLAB (Space Rendezvous Laboratory) organized a competition that brought together 48 participants. Each team proposed a solution that is based at least in part on deep neural networks, now a dominant technique \cite{kisantal_satellite_2020}. The competition was based on the SPEED (Spacecraft PosE Estimation Dataset) dataset introduced earlier by Sharma \textit{et. al} \cite{sharma_pose_2019}. The dataset contains 12000 synthetic images of the Tango satellite to train the models; 2998 synthetic images on which the participants were ranked (synthetic test set); and 300 real images, which allow characterizing the generalization capacity of the proposed models (real test set). A post-mortem webpage dedicated to model predictions evaluation is still available as the labels of the test sets are not provided \cite{noauthor_kelvins_nodate-1}.


Sharma \textit{et al.} \cite{sharma_pose_2019}, Chan \textit{et al.} \cite{chen_satellite_2019} and the EPFL CVLab team used a 3-step process to estimate spacecraft pose \cite{kisantal_satellite_2020}. First, they use an object detection CNN to determine the region of interest and crop the input image. Then, another CNN regress keypoints. Finally, they solve Pose estimation using an off-the-shelf Perspective-n-Point (PnP) solver. Black \textit{et al.} follow the same 3-step process, but instead of using large CNNs, they use a MobileNet-v2 as keypoint regression network. Nevertheless, their method uses a complex pipeline in which the MobileNet-v2 CNN only represents 20.4\% of the inference execution time\cite{black_real-time_2021}. Moreover, using keypoints limits the method to known spacecraft. 


Proença \textit{et al.} proposed URSONet: a straightforward way to solve SPE by regressing position and orientation using a single CNN \cite{proenca_deep_2020}. It allows to directly optimize ESA competition metrics \cite{kisantal_satellite_2020}. They also proposed to deal with the orientation estimation as a soft classification task which significantly improves the results. The orientation is encoded as a Gaussian random variable in a discrete output space so that the CNN learns to predict a mass density function \cite{proenca_deep_2020}. Then, they used a softmax function and the quaternion averaging technique to predict the orientation \cite{markley_averaging_2007}. URSONet stands out for its generalization capabilities as Proença \textit{et al.} obtained a good score on both synthetic and real test sets. In addition, as they do not rely on keypoints, their method would be able to generalize to objects with unknown geometry using SLAM \cite{proenca_deep_2020}. However, it comes at the cost of a tremendous amount of parameters (500 million) and high computational complexity as they use an ensemble method based on three ResNet-101 CNNs \cite{proenca_deep_2020}.




In the spirit of MobileNet proposed by Google \cite{howard_mobilenets_2017}, we propose Mobile-URSONet: a spacecraft pose estimation convolutional neural network adapted to spacecraft onboard computers. Our lightest model has 178 times fewer parameters while degrading accuracy by no more than four times compared to URSONet.

The outline of this paper is as follows: Section \ref{sec:methodology} explains the methodology we adopted to optimize URSONet for embedded systems. Section \ref{sec:experiments} presents the experimental results obtained, and section \ref{sec:conslusion} concludes the paper.

%% file: 3-methodology.tex
\section{PROPOSED METHOD}\label{sec:methodology}



Our analysis is based on the ESA competition evaluation metrics \cite{kisantal_satellite_2020} which includes: the mean absolute position error $e_t$ (in meters), the mean absolute orientation error $e_q$ (in degrees), and the mean ESA score $E$ (lower is better) which is evaluated on both the synthetic ($E_{syn}$) and real test sets ($E_{real}$).











\subsection{Generalization metric}

Many solutions proposed during the ESA competition do not obtain as good a score on the real images as on synthetic images. Kisantal \textit {et al.} explains that it comes from the change in statistical distribution between the synthetic and the real images \cite {kisantal_satellite_2020}. However, they do not explain the huge differences in generalization between the different models. We propose a generalization metric to characterize these differences: $G_ {factor}$. It represents the ratio between the mean ESA score obtained on the real and synthetic test sets:

\begin{equation}
    G_{factor} = \frac{E_{real}}{E_{syn}}
\end{equation}


Table \ref{tab:speed_gfactor} presents the results and generalization factors of the first four participants of the ESA competition \cite {kisantal_satellite_2020} (including the baseline solution of Stanford SLAB \cite {sharma_pose_2019}), and the latest work published on the domain by Black \textit {et al.} \cite {black_real-time_2021}. We observe a wide variability in generalization factors: from 2.7 for the third in the competition (Proença \textit{et al.}) to 39.9 for the winners of the competition (Chen \textit{et al.}). Beyond solving the SPE task using a single CNN (without PnP methods), which is a path to pose estimation of unknown objects, Proença \textit{et al.} solution offers the best generalization factor. As robustness to changes in distribution is a key criterion when integrating such networks into embedded systems, we base our work on that of Proença \textit{et al.}

\begin{table}[h]
    \centering
    \caption{ESA score on the test sets and corresponding generalization factor}
    \begin{tabular}{|l|c|c|c|}
        \hline
        Participants                & $E_{synth}$ & $E_{real}$ & $G_{factor}$ \\
        \hline
        Chen \textit{et al.} \cite{chen_satellite_2019}        & 0.0094    & 0.3752     & 39.9          \\
        EPFL\_cvlab                                            & 0.0215    & 0.1140     & 5.3           \\
        Black \textit{et al.} \cite{black_real-time_2021}      & 0.0409    & 0.2918     & 7.13          \\
        Proença \textit{et al.} \cite{proenca_deep_2020}       & 0.0571    & 0.1555     & 2.7           \\
        Sharma \textit{et al.} \cite{sharma_pose_2019}         & 0.0626    & 0.3951     & 6.3           \\
        \hline
    \end{tabular}
    \label{tab:speed_gfactor}
\end{table}

\subsection{Neural network architecture}

\begin{figure}[h]
    \includegraphics[scale=0.215]{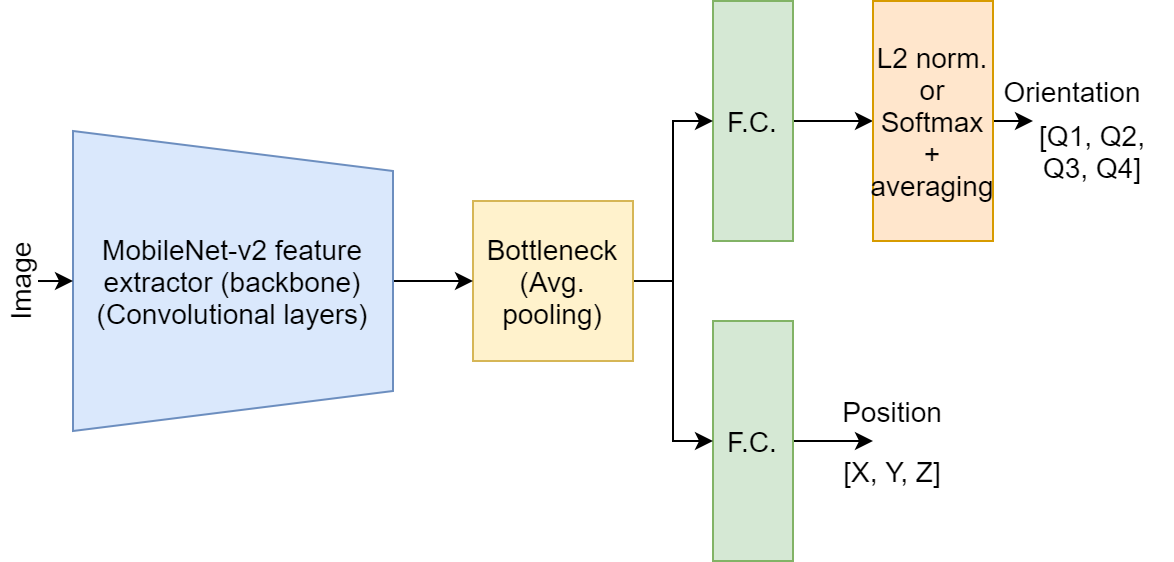}
    \centering
    \caption{Mobile-URSONet architecture}
    \label{fig:cnn_architecture}
\end{figure}

To ensure a good ranking in the ESA competition, Proença \textit{et. al.} focused on minimizing the ESA score regardless of the complexity of their model. We focused on the trade-off between the ESA score, the number of parameters, and the computational complexity of our model. Figure \ref{fig:cnn_architecture} shows an outline of Mobile-URSONet, the neural network we propose in this work. In the following paragraphs, we will explain in detail our architectural choices.

The first component of a CNN is the backbone that extracts features from the input image (Fig. \ref{fig:cnn_architecture}). Proença \textit{et al.} used a ResNet-101 backbone, a CNN that achieves 77\% top-1 accuracy on the standard ImageNet benchmark. We use a MobileNet-v2 backbone that achieves only 72\% top-1 accuracy on ImageNet but has 13 times fewer parameters. Bianco \textit{et. al.} show that MobileNet-V2 has a top-1 accuracy density (\textit{i.e.} accuracy per million parameters) more than 10 times better than the ResNet-101 used by Proença \textit{et al.} \cite{bianco_benchmark_2018}. They also show that MobileNet-v2 inference requires only 0.3 GFLOPs (floating-point operations), while a ResNet-101 inference requires almost 8 GFLOPs. The computational complexity of our backbone is 26 times less than the original URSONet.


Moreover, MobileNet-v2 is designed to run in real-time on smartphone ARM processors \cite{sandler_mobilenetv2_2018}. Table \ref{tab:latency_mobilenet} shows that it also run on weaker ARM processors now used in the space domain. For instance, the Eye-Sat nanosatellite embeds a Zynq 7030 MPSoC (Multiprocessor System on a Chip) developed by Xilinx, which has two ARM A-9 cores \cite{apper_eye-sat_nodate}. More recent projects consider using MPSoC featuring four ARM A-53 cores \cite{perez_run-time_2020} \cite{fuchs_fault-tolerant_2019}.

\begin{table}[h]
    \centering
    \caption{Inference latency of MobileNet-v2 on various ARM processors \cite{sandler_mobilenetv2_2018} \cite{moreau_hardware-software_2019}}
    \begin{tabular}{|l|c|}
        \hline
        Processor (compiler)               & Latency \\
        \hline
        Qualcomm ARM A-72 (TF-Lite)        & 75 ms \\
        Xilinx ARM A-53 (TVM)              & 132 ms \\
        Xilinx ARM A-9 (TVM)               & 265 ms \\
        \hline
    \end{tabular}
    \label{tab:latency_mobilenet}
\end{table}


The second part of the neural network is the bottleneck (Fig. \ref{fig:cnn_architecture}). Many standard CNN architectures uses average pooling such as MobileNets \cite{howard_mobilenets_2017} \cite{sandler_mobilenetv2_2018} and ResNets \cite{he_deep_2015}. It aims to merge semantically similar features \cite{lecun_deep_2015} by reducing the spatial dimension of the feature maps. Thus, it reduces the number of parameters and computational complexity of the subsequent layers. Proença \textit{et al.} replaced the standard average pooling layer by a 3x3 convolution layer with a stride of 2 \cite{proenca_deep_2020}. They show that increasing the number of feature maps in the bottleneck layer decrease the position and orientation error. We summarize their results in table \ref{tab:bottleneck_effect}. The first line of the table (\textit{i.e.} the configuration with eight feature maps) has the same amount of parameters as an average pooling configuration. The table shows that multiplying the number of parameters by six only leads to a 3 degrees improvement in orientation error and 0.24 meters improvement in position error. In our opinion, it is not a suitable trade-off for an embedded system. That is why we kept the original average pooling layer.

\begin{table}[h]
    \centering
    \caption{Bottleneck size \textit{vs.} number of parameters, orientation and position error on URSONet \cite{proenca_deep_2020}}
    \begin{tabular}{|c|c|c|c|}
        \hline
        \# feature maps  & \# params (M) & Ori err. (°) & Pos err. (m)\\
        \hline
        8                & 40            & 10.2          & 0.72        \\
        128              & 80            & 7.8           & 0.54         \\
        512              & 240           & 7.2           & 0.48          \\
        \hline
    \end{tabular}
    \label{tab:bottleneck_effect}
\end{table}


Moreover, the number of parameters of the neural network is agnostic to input image resolution thanks to the average pooling layer. Proença \textit{et al.} show that orientation estimation is sensitive to image resolution (\textit{i.e.} increasing image resolution improves accuracy). Thus, increasing image resolution is a more efficient solution to improve accuracy than removing the bottleneck as the power consumption highly depends on memory accesses \cite{horowitz_11_2014}.


The last part of the neural network is the two branches (Fig. \ref{fig:cnn_architecture}): the first part estimates orientation, while the other estimates position. Proença \textit{et al.} use two fully connected layers. We use a single fully connected layer to minimize the number of parameters. Our position branch has only 3843 parameters. The number of parameters of our orientation branch depends on the configuration: in regression, it has 5124 parameters; in soft classification, it has between 0.6 and 5 million parameters. It is 39 to 325 times fewer parameters than Proença \textit{et al.}, which has approximately 195 million parameters in their branches.

\subsection{Loss functions}



Target's position estimation is an easy task solved using direct regression. It allows using ESA metrics as loss functions, as proposed by Proença \textit{et al.}. The same cannot apply to orientation estimation, which explains why we try two methods (\textit{i.e.} regression and soft classification). In ESA metrics, the position error depends on the distance from the target spacecraft; the neural network is increasingly penalized as the target spacecraft is closer. However, this is not the case with the orientation in ESA metrics. We propose a variant of the loss function when orientation estimation is considered a regression task:

\begin{equation}
    L_{ori} = \frac{\arccos{ | \textbf{q} \cdot \hat{\textbf{q}} |}}{||\textbf{t}||_2}
    \label{eq:l_ori_modif}
\end{equation}


In our experiments, it does not lead to significant improvements compared to the regression loss function proposed by Proença \textit{et. al}. Thus, we also estimate orientation using soft classification as Proença \textit{et. al.} does. The associated loss function is a standard negative log-likelihood \cite{proenca_deep_2020}.

%% file: 4-experiments.tex
\section{EXPERIMENTS}\label{sec:experiments}

\subsection{Implementation and training details}


Networks are trained on one Nvidia Tesla P100, using Pytorch on the SPEED dataset. We use the MobileNet-v2 backbone pre-trained on ImageNet to speed up training. We reserve 15\% of the training SPEED dataset as a validation set. Parameters are updated using the SGD algorithm with a \textit{momentum} of 0.9. We use a batch size of 32 images resized to 384 * 240 pixels. The learning rate starts at 0.01 for the first 30 epochs. Then it is decayed to 0.001 for the following 15 epochs. It finishes at 0.0001 for the last five epochs. We employ data augmentation on the training set using OpenCV to rotate the camera across the roll axis for half images with a maximum magnitude of 25°. We also use Pytorch transformations to add a Gaussian blur and randomly change the brightness, contrast, saturation, and hue of training images. When using soft classification, we have set $\Delta$ which controls the Gaussian width at 3 to act as a regularizer. The number of bins per dimension varies between 8 and 32. All hyper-parameters are tuned on the validation set. Our code is available at \cite{noauthor_mobile-ursonet_nodate}.

\subsection{Results}

Orientation estimation through regression is done with very few parameters but leads to an average error of 32° on the validation set. Orientation estimation through soft classification implies many more parameters in the orientation branch: it depends on the cube of the number of bins per dimension (we encode rotations as three Euler angles and then convert it to quaternions). However, using soft classification improves the orientation error by a factor of three to five. Table \ref{tab:n_bins_ori_error} shows the results we obtain, with 8 to 32 bins per dimension. The 16-bins model has 2.6 times more parameters than the 8-bins model and an orientation error divided by 2. However, the 32-bins model has six times more parameters than the 16-bins model, with no significant improvement on the orientation error. The 24-bins model also does not bring improvements. It demonstrates a saturation effect on orientation error when increasing the number of bins per dimension. In addition, we notice that the 32-bins model is much more prone to overfitting. Based on these results, we believe that going beyond 16 bins per dimension is not worth it as we focus on the trade-off between the orientation error and the number of parameters of the CNN.

\begin{table}[h]
    \centering
    \caption{Effect of the number of bins per dimension on orientation error and the number of parameters}
    \begin{tabular}{|c|c|c|c|}
        \hline
        \# bins per dim. & \# params (M) & $e_q$ train (°) & $e_q$ valid (°) \\
        \hline
        8                & 2.8           & 9.74           & 11.3          \\
        12               & 4.4           & 5.69           & 7.43          \\
        16               & 7.4           & 4.5            & 6.29           \\
        24               & 19.9          & 4.18           & 6.12           \\
        32               & 44.2          & 4.92           & 7.29           \\
        \hline
    \end{tabular}
    \label{tab:n_bins_ori_error}
\end{table}

Table \ref{tab:errors_compared_urso} shows the orientation and position error of our selected models (from 8 to 16 bins) compared to \cite{proenca_deep_2020}. As we said before, target position estimation is an easy task solved using direct regression. Position error is the same in all our experiments. It seems to depend only on the number of parameters of the backbone, as increasing the number of parameters in the position branch only increases overfitting. Our backbone uses 13 times fewer parameters than Proença's backbone while degrading position error by no more than three times. Using soft classification, orientation error highly depends on the number of parameters of the orientation branch. Proença \textit{et al.} used between 24 and 64 bins per dimension but only published the orientation error for their 24-bins model. The high number of parameters of URSONet causes overfitting that Proença \textit{et al.} mitigates using a data augmentation strategy more refined than ours. It explains why they achieve a better orientation error of 4.0°, while our best model achieves only 6.3° orientation error.

\begin{table}[h]
    \centering
    \caption{Orientation and position error on validation set compared to Proença \textit{et al.}}
    \begin{tabular}{|l|c|c|c|}
        \hline
        \# bins per dim. & $e_q$ (°) & $e_t$ (m) \\
        \hline
        Ours (8 bins)    & 11.3           & 0.54          \\
        Ours (12 bins)   & 7.43           & 0.51          \\
        Ours (16 bins)   & 6.29           & 0.56           \\
        Proença \textit{et al.} (24 bins) \cite{proenca_deep_2020}
                         & 4.0            & 0.17           \\
        \hline
    \end{tabular}
    \label{tab:errors_compared_urso}
\end{table}

\begin{figure}[h]
    \includegraphics[scale=0.42]{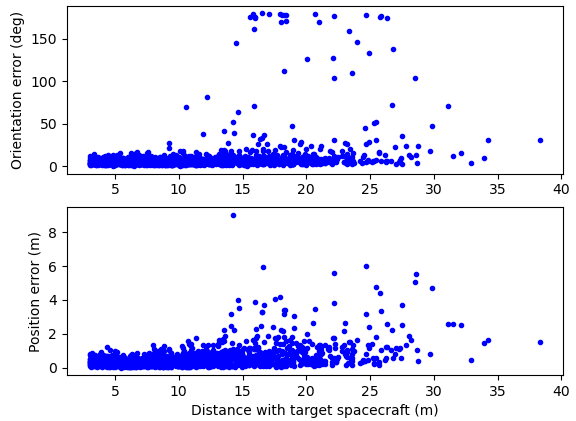}
    \centering
    \caption{Position and orientation error by distance for our 12-bins model}
    \label{fig:distance_error}
\end{figure}

Figure \ref{fig:distance_error} shows the position and the orientation error of our 12-bins model on the validation set as a function of distance to the target. We see that the position error highly depends on the distance to the target satellite. We observe the same property for the orientation error. It is surprising as the loss function we use does not involve the distance with the target spacecraft (in soft classification configuration). High position and orientation errors appear when the target spacecraft is more than 10 meters apart from the camera. The number of outliers is small enough and occurs only when the target spacecraft is more than 15 meters apart from the camera. The closer the target spacecraft is, the more confidence we can have in the predictions of our model. It is a crucial property for autonomous docking or debris removal applications. It also offers an avenue to reduce both the position and the orientation error: zooming and cropping the image around the target instead of resizing the whole image as we do now.

Table \ref{tab:final_results} summarizes the results of our models compared to our competitors. We propose the most lightweight spacecraft pose estimation models: ranging from 2.2 to 7.4 million parameters while keeping a good score on the synthetic test set and a good generalization factor. Our 8-bins model has 178 times fewer parameters while degrading the ESA score by no more than four times compared to URSONet. It has an accuracy density (\textit{i.e.} ESA score per parameters) 139 times higher than the original URSONet. However, we notice that increasing the number of bins per dimension leads the model to overfit the synthetic images. Proença \textit{et al.} demonstrated that a 2.7 generalization factor is achievable while using 24 to 64 bins per orientation dimension. We believe we still have some margin to improve the generalization capabilities of our models by using a more advanced data augmentation technique and by investing more effort in hyper-parameter tuning.

\begin{table}[h]
    \centering
    \caption{ESA score on test set and generalization factor}
    \begin{tabular}{|l|c|c|c|c|}
        \hline
        Participants                                           & \# params (M)  & $E_{synth}$ & $E_{real}$ & $G_{factor}$ \\
        \hline
        Black \textit{et al.} \cite{black_real-time_2021}      & 6.9    & 0.0409    & 0.2918     & 7.13          \\
        Sharma \textit{et al.} \cite{sharma_pose_2019}         & 11.2   & 0.0626    & 0.3951     & 6.31           \\
        Proença \textit{et al.} \cite{proenca_deep_2020}       & 500    & 0.0571    & 0.1555     & 2.72           \\
        ours (regression)                                      & 2.2    & 0.6160    & 0.7997     & 1.30          \\
        ours (8 bins)                                  & 2.8    & 0.2520    & 0.7868     & 3.12          \\
        ours (12 bins)                                 & 4.4    & 0.2104    & 1.2231     & 5.81          \\
        ours (16 bins)                                         & 7.4    & 0.1947    & 1.2074     & 6.20          \\
        \hline
    \end{tabular}
    \label{tab:final_results}
\end{table}
\subsection{Future work}


Future works will explore in-depth embeddability by using quantization and pruning. It will further optimize the memory footprint of the parameters and the inference computational complexity of Mobile-URSONet. We plan to deploy these models on promising commercial chips for future satellite onboard computers, such as the Xilinx MPSoCs featuring ARM-A53 cores and programmable logic.

%% file: 5-conclusions.tex
\section{CONCLUSIONS}\label{sec:conslusion}

In this paper, we analyzed URSONet, a popular neural network used for spacecraft pose estimation that seems hardly compatible with the constraints of onboard spacecraft computers. We found that three architectural choices have a dominant effect on both the number of parameters and the computational complexity of the CNN: the backbone, the bottleneck size, and the number of bins per dimension while predicting orientation using soft classification. By analyzing trade-offs for each of these three architectural choices, we were able to propose Mobile-URSONet, a mobile version of URSONet in the spirit of Google MobileNets. We showed that Mobile-URSONet achieves accuracy close to URSONet, while keeping the number of parameters and computational complexity compatible with the constraints of spacecraft onboard computers.

%% file: 7-acknowledgment.tex
\section{ACKNOWLEDGMENTS}

The authors thank the Canadian Space Agency, MITACS and Space Codesign Systems for their financial contributions.

%% file: root.bbl
\begin{thebibliography}{10}
\providecommand{\url}[1]{#1}
\csname url@samestyle\endcsname
\providecommand{\newblock}{\relax}
\providecommand{\bibinfo}[2]{#2}
\providecommand{\BIBentrySTDinterwordspacing}{\spaceskip=0pt\relax}
\providecommand{\BIBentryALTinterwordstretchfactor}{4}
\providecommand{\BIBentryALTinterwordspacing}{\spaceskip=\fontdimen2\font plus
\BIBentryALTinterwordstretchfactor\fontdimen3\font minus
  \fontdimen4\font\relax}
\providecommand{\BIBforeignlanguage}[2]{{%
\expandafter\ifx\csname l@#1\endcsname\relax
\typeout{** WARNING: IEEEtran.bst: No hyphenation pattern has been}%
\typeout{** loaded for the language `#1'. Using the pattern for}%
\typeout{** the default language instead.}%
\else
\language=\csname l@#1\endcsname
\fi
#2}}
\providecommand{\BIBdecl}{\relax}
\BIBdecl

\bibitem{proenca_deep_2020}
P.~F. Proença and Y.~Gao, ``Deep learning for spacecraft pose estimation from
  photorealistic rendering,'' in \emph{2020 {IEEE} {International} {Conference}
  on {Robotics} and {Automation} ({ICRA})}.\hskip 1em plus 0.5em minus
  0.4em\relax IEEE, 2020, pp. 6007--6013.

\bibitem{noauthor_kelvins_nodate}
\BIBentryALTinterwordspacing
``Kelvins - {Pose} {Estimation} {Challenge}.'' [Online]. Available:
  \url{https://kelvins.esa.int/satellite-pose-estimation-challenge/home/}
\BIBentrySTDinterwordspacing

\bibitem{noauthor_esa_nodate}
\BIBentryALTinterwordspacing
``\BIBforeignlanguage{en}{{ESA} commissions world’s first space debris
  removal}.'' [Online]. Available:
  \url{https://www.esa.int/Safety_Security/Clean_Space/ESA_commissions_world_s_first_space_debris_removal}
\BIBentrySTDinterwordspacing

\bibitem{sharma_pose_2018}
S.~Sharma, C.~Beierle, and S.~D'Amico, ``Pose estimation for non-cooperative
  spacecraft rendezvous using convolutional neural networks,'' in \emph{2018
  {IEEE} {Aerospace} {Conference}}, Mar. 2018, pp. 1--12.

\bibitem{kisantal_satellite_2020}
\BIBentryALTinterwordspacing
M.~Kisantal, S.~Sharma, T.~H. Park, D.~Izzo, M.~Märtens, and S.~D'Amico,
  ``Satellite {Pose} {Estimation} {Challenge}: {Dataset}, {Competition}
  {Design} and {Results},'' \emph{arXiv:1911.02050 [cs]}, Apr. 2020, arXiv:
  1911.02050. [Online]. Available: \url{http://arxiv.org/abs/1911.02050}
\BIBentrySTDinterwordspacing

\bibitem{sharma_pose_2019}
\BIBentryALTinterwordspacing
S.~Sharma and S.~D'Amico, ``Pose {Estimation} for {Non}-{Cooperative}
  {Rendezvous} {Using} {Neural} {Networks},'' \emph{arXiv:1906.09868 [cs]},
  Jun. 2019, arXiv: 1906.09868. [Online]. Available:
  \url{http://arxiv.org/abs/1906.09868}
\BIBentrySTDinterwordspacing

\bibitem{noauthor_kelvins_nodate-1}
\BIBentryALTinterwordspacing
``Kelvins - {Pose} {Estimation} {Challenge} post mortem - {Home}.'' [Online].
  Available:
  \url{https://kelvins.esa.int/satellite-pose-estimation-challenge/leaderboard/post-mortem-leaderboard}
\BIBentrySTDinterwordspacing

\bibitem{chen_satellite_2019}
\BIBentryALTinterwordspacing
B.~Chen, J.~Cao, A.~Parra, and T.-J. Chin, ``Satellite {Pose} {Estimation} with
  {Deep} {Landmark} {Regression} and {Nonlinear} {Pose} {Refinement},''
  \emph{arXiv:1908.11542 [cs]}, Aug. 2019, arXiv: 1908.11542. [Online].
  Available: \url{http://arxiv.org/abs/1908.11542}
\BIBentrySTDinterwordspacing

\bibitem{black_real-time_2021}
\BIBentryALTinterwordspacing
K.~Black, S.~Shankar, D.~Fonseka, J.~Deutsch, A.~Dhir, and M.~R. Akella,
  ``Real-{Time}, {Flight}-{Ready}, {Non}-{Cooperative} {Spacecraft} {Pose}
  {Estimation} {Using} {Monocular} {Imagery},'' \emph{arXiv:2101.09553 [cs]},
  Jan. 2021, arXiv: 2101.09553. [Online]. Available:
  \url{http://arxiv.org/abs/2101.09553}
\BIBentrySTDinterwordspacing

\bibitem{markley_averaging_2007}
\BIBentryALTinterwordspacing
F.~L. Markley, Y.~Cheng, J.~L. Crassidis, and Y.~Oshman,
  ``\BIBforeignlanguage{en}{Averaging {Quaternions}},''
  \emph{\BIBforeignlanguage{en}{Journal of Guidance, Control, and Dynamics}},
  vol.~30, no.~4, pp. 1193--1197, Jul. 2007. [Online]. Available:
  \url{https://arc.aiaa.org/doi/10.2514/1.28949}
\BIBentrySTDinterwordspacing

\bibitem{howard_mobilenets_2017}
\BIBentryALTinterwordspacing
A.~G. Howard, M.~Zhu, B.~Chen, D.~Kalenichenko, W.~Wang, T.~Weyand,
  M.~Andreetto, and H.~Adam, ``{MobileNets}: {Efficient} {Convolutional}
  {Neural} {Networks} for {Mobile} {Vision} {Applications},''
  \emph{arXiv:1704.04861 [cs]}, Apr. 2017, arXiv: 1704.04861. [Online].
  Available: \url{http://arxiv.org/abs/1704.04861}
\BIBentrySTDinterwordspacing

\bibitem{bianco_benchmark_2018}
\BIBentryALTinterwordspacing
S.~Bianco, R.~Cadene, L.~Celona, and P.~Napoletano, ``Benchmark {Analysis} of
  {Representative} {Deep} {Neural} {Network} {Architectures},'' \emph{IEEE
  Access}, vol.~6, pp. 64\,270--64\,277, 2018, arXiv: 1810.00736. [Online].
  Available: \url{http://arxiv.org/abs/1810.00736}
\BIBentrySTDinterwordspacing

\bibitem{sandler_mobilenetv2_2018}
\BIBentryALTinterwordspacing
M.~Sandler, A.~Howard, M.~Zhu, A.~Zhmoginov, and L.-C. Chen,
  ``\BIBforeignlanguage{en}{{MobileNetV2}: {Inverted} {Residuals} and {Linear}
  {Bottlenecks}},'' in \emph{\BIBforeignlanguage{en}{2018 {IEEE}/{CVF}
  {Conference} on {Computer} {Vision} and {Pattern} {Recognition}}}.\hskip 1em
  plus 0.5em minus 0.4em\relax Salt Lake City, UT: IEEE, Jun. 2018, pp.
  4510--4520. [Online]. Available:
  \url{https://ieeexplore.ieee.org/document/8578572/}
\BIBentrySTDinterwordspacing

\bibitem{apper_eye-sat_nodate}
F.~Apper, A.~Ressouche, N.~Humeau, M.~Vuillemin, A.~Gaboriaud, F.~Viaud, and
  M.~Couture, ``\BIBforeignlanguage{en}{Eye-{Sat}: {A} {3U} student {CubeSat}
  from {CNES} packed with technology},'' p.~10.

\bibitem{perez_run-time_2020}
A.~Pérez, A.~Rodríguez, A.~Otero, D.~G. Arjona, {\'A}.~Jiménez-Peralo,
  M.~{\'A}. Verdugo, and E.~De~La~Torre, ``Run-{Time} {Reconfigurable}
  {MPSoC}-{Based} {On}-{Board} {Processor} for {Vision}-{Based} {Space}
  {Navigation},'' \emph{IEEE Access}, vol.~8, pp. 59\,891--59\,905, 2020,
  conference Name: IEEE Access.

\bibitem{fuchs_fault-tolerant_2019}
C.~M. Fuchs, P.~Chou, X.~Wen, N.~M. Murillo, G.~Furano, S.~Holst,
  A.~Tavoularis, S.-K. Lu, A.~Plaat, and K.~Marinis, ``A {Fault}-{Tolerant}
  {MPSoC} {For} {CubeSats},'' in \emph{2019 {IEEE} {International} {Symposium}
  on {Defect} and {Fault} {Tolerance} in {VLSI} and {Nanotechnology} {Systems}
  ({DFT})}, Oct. 2019, pp. 1--6, iSSN: 2377-7966.

\bibitem{moreau_hardware-software_2019}
\BIBentryALTinterwordspacing
T.~Moreau, T.~Chen, L.~Vega, J.~Roesch, E.~Yan, L.~Zheng, J.~Fromm, Z.~Jiang,
  L.~Ceze, C.~Guestrin, and A.~Krishnamurthy, ``A {Hardware}-{Software}
  {Blueprint} for {Flexible} {Deep} {Learning} {Specialization},''
  \emph{arXiv:1807.04188 [cs, stat]}, Apr. 2019, arXiv: 1807.04188. [Online].
  Available: \url{http://arxiv.org/abs/1807.04188}
\BIBentrySTDinterwordspacing

\bibitem{he_deep_2015}
\BIBentryALTinterwordspacing
K.~He, X.~Zhang, S.~Ren, and J.~Sun, ``Deep {Residual} {Learning} for {Image}
  {Recognition},'' \emph{arXiv:1512.03385 [cs]}, Dec. 2015, arXiv: 1512.03385.
  [Online]. Available: \url{http://arxiv.org/abs/1512.03385}
\BIBentrySTDinterwordspacing

\bibitem{lecun_deep_2015}
\BIBentryALTinterwordspacing
Y.~LeCun, Y.~Bengio, and G.~Hinton, ``\BIBforeignlanguage{en}{Deep learning},''
  \emph{\BIBforeignlanguage{en}{Nature}}, vol. 521, no. 7553, pp. 436--444, May
  2015, number: 7553 Publisher: Nature Publishing Group. [Online]. Available:
  \url{https://www.nature.com/articles/nature14539}
\BIBentrySTDinterwordspacing

\bibitem{horowitz_11_2014}
M.~Horowitz, ``1.1 {Computing}'s energy problem (and what we can do about
  it),'' in \emph{2014 {IEEE} {International} {Solid}-{State} {Circuits}
  {Conference} {Digest} of {Technical} {Papers} ({ISSCC})}, Feb. 2014, pp.
  10--14, iSSN: 2376-8606.

\bibitem{noauthor_mobile-ursonet_nodate}
\BIBentryALTinterwordspacing
``Mobile-{URSONet} code ({GitHub}).'' [Online]. Available:
  \url{https://github.com/possoj/Mobile-URSONet}
\BIBentrySTDinterwordspacing

\end{thebibliography}
